\PassOptionsToPackage{table}{xcolor}
\documentclass{article} 
\usepackage{iclr2026_conference,times}

\usepackage{enumitem}
\usepackage{url}
\usepackage[utf8]{inputenc} 
\usepackage[T1]{fontenc}    
\usepackage{url}            
\usepackage{booktabs}       
\usepackage{amsfonts}       
\usepackage{nicefrac}       
\usepackage{microtype}      
\usepackage[table]{xcolor}         
\usepackage{graphicx}
\usepackage{subfigure}
\usepackage{mathtools}
\usepackage{amsthm}
\usepackage{amsmath}
\usepackage{amssymb}
\usepackage{extarrows}
\usepackage{diagbox}
\usepackage{microtype}
\usepackage{multirow}
\usepackage{makecell}

\newcommand{\method}{\texttt{HWM}}

\usepackage{amsmath,amsfonts,bm}

\def\eqref#1{equation~\ref{#1}}

\def\1{\bm{1}}

\DeclareMathAlphabet{\mathsfit}{\encodingdefault}{\sfdefault}{m}{sl}
\SetMathAlphabet{\mathsfit}{bold}{\encodingdefault}{\sfdefault}{bx}{n}

\definecolor{text}{rgb}{0.4,0.1,0.4}

\usepackage{algorithm}
\usepackage{algorithmic}

\title{Structured Latent Dynamics for Long-Horizon Embodied World Models}

\author{
Kevin Chen*\quad 
Yu-Wei Zhan*\quad 
Haoran Liang\quad 
Sen Cui\textsuperscript{\ddag,\dag}\quad 
Changshui Zhang\textsuperscript{\dag}\\
Tsinghua University\\
\textsuperscript{*}Equal contribution. \quad \textsuperscript{\ddag}Project leader. \quad \textsuperscript{\dag}Corresponding authors. 
}

\makeatletter
\renewcommand{\headrulewidth}{0pt}  
\makeatother
\iclrfinalcopy 
\usepackage{hyperref}
\begin{document}

\maketitle

\begin{abstract}
Reliable long-horizon prediction remains a central challenge for embodied world models. Although recent generative models can synthesize visually plausible action-conditioned futures, their latent dynamics are typically learned through unconstrained transition modules, which may gradually deviate from coherent physical evolution as predictions are recursively rolled forward. We investigate whether explicitly structuring latent state evolution can improve the stability and utility of such imagined trajectories. To this end, we introduce \method{}, a structured dynamics framework that represents backbone latents in a learned phase space consisting of generalized coordinates and momenta. Future states are propagated through dissipative Hamiltonian dynamics and mapped back to the original latent representation through a learnable inverse interface. The resulting formulation can be integrated into existing world-model architectures without redesigning their perception or decoding components, while multi-step transition supervision encourages the structured rollout to remain aligned with the latent trajectories induced by real observations. Controlled simulation experiments show that \method{} produces more accurate estimates of velocity, acceleration, and jerk and substantially reduces long-horizon energy drift relative to unstructured latent dynamics. We further evaluate the approach across three distinct embodied world-model settings, covering single-arm multi-view prediction with Ctrl-World, dual-arm single-view prediction with EVAC, and task-oriented single-arm prediction with Cosmos-Predict2.5-2B. On RoboCasa, incorporating \method{} into Cosmos-Predict2.5-2B yields a 2.7\% relative improvement in task success. On real-world AgiBot World trajectories, it improves nDTW by 9.7\% with Ctrl-World and Logic by 8.4\% with EVAC. Across these settings, the results indicate that introducing explicit dynamical structure into latent evolution can improve both long-horizon predictive stability and the downstream usefulness of embodied world-model rollouts. The project page is available at \url{https://kevinc-lab.github.io/Structured-Latent-Dynamics/}.
\end{abstract}
\vspace{-0.5cm}

\begin{figure}[H]
    \centering
    \includegraphics[width=\linewidth]{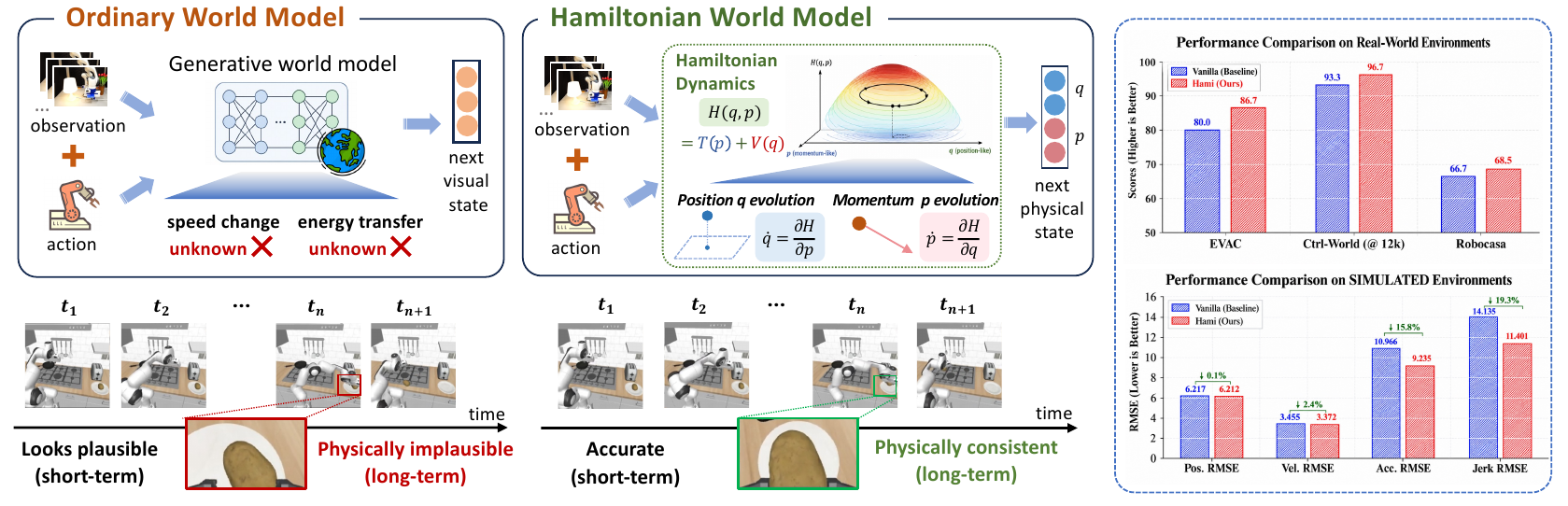}
    \vspace{-0.6cm}
    \caption{Comparison between ordinary generative world models and our Hamiltonian World Model.}
    \vspace{-0.3cm}
    \label{fig:intro}
\end{figure}

\section{Introduction}

World models have long been studied as internal simulators for sequential decision making, enabling planning, counterfactual reasoning, and data-efficient policy improvement \citep{sutton1991dyna, ha2018worldmodels, hafner2019planet, hafner2020dreamer, hafner2023dreamerv3}. Recent advances in generative video modeling have pushed this paradigm into embodied manipulation, where a model is expected to predict action-conditioned futures for robotic interaction \citep{hu2023gaia, bruce2024genie, seo2023multiviewmaskedworldmodels, zhou2024robodreamerlearningcompositionalworld, zhu2025irasimfinegrainedworldmodel, guo2026ctrlworldcontrollablegenerativeworld, nvidia2025cosmosworldfoundationmodel}. In this setting, however, the value of a world model is not determined only by visual realism. A predicted future must also remain physically coherent enough to support downstream policy evaluation and long-horizon execution.

This requirement exposes a central limitation of many modern embodied world models. Although current architectures can generate visually plausible short-term videos, their latent transitions are often implemented as generic sequence or generative modules. Such transitions can fit local appearance statistics and short-range correlations, but they do not explicitly encode the dynamical structure that governs physical interaction. As a result, small one-step errors may compound into long-horizon physical drift: objects may penetrate, contact timing may become inconsistent, and action effects may gradually decouple from the commanded motion. For an embodied agent, these failures directly weaken the usefulness of imagined rollouts as substrates for planning and control.

Existing approaches address this issue from two directions. One line of work scales generative world models and improves controllability, view consistency, and open-loop prediction quality \citep{hu2023gaia, bruce2024genie, zhou2024robodreamerlearningcompositionalworld, zhu2025irasimfinegrainedworldmodel, guo2026ctrlworldcontrollablegenerativeworld, nvidia2025cosmosworldfoundationmodel}. These models provide strong perceptual and rendering capabilities, but the physical regularities of manipulation are still mostly learned implicitlym, shown in Figure~\ref{fig:intro}. Another line introduces physical structure into learned dynamics. For example, Symplectic ODE-Nets show that structure-preserving formulations can improve stability and extrapolation in dynamical systems \citep{greydanus2019hamiltonian, toth2020hamiltonian, zhong2020symplectic}. Yet the embodied video world-model setting requires a different design target: a transition mechanism that can be inserted into modern latent visual world models while preserving their existing observation, tokenization, and rendering interfaces.

Our key conceptual claim is that physical consistency should not be treated as an auxiliary property imposed after a generic transition has been learned. Instead, the transition law itself should be physically structured. In other words, the central modeling object should shift from ``a neural sequence model that predicts the next latent'' to ``a learned physical evolution rule in latent phase space.'' This view is natural for embodied manipulation: a robot does not merely observe image sequences, but interacts with a dynamical system through forces, contacts, inertia, and dissipation. A world model that aims to support action should represent not only appearance evolution, but also the latent physical state whose evolution determines whether an imagined future remains executable.

Motivated by this view, we propose Hamiltonian World Model (\method{}), a physically structured latent transition framework for embodied world modeling. \method{} maps backbone latents into learned generalized coordinates and momenta, which evolve under a dissipative Hamiltonian system. The Hamiltonian models the energy landscape, while damping accounts for non-conservative effects such as friction. A semi-implicit Euler update reduces numerical drift, and the predicted phase states are mapped back to backbone latents for decoding. This design structures long-horizon dynamics while preserving the backbone's perceptual and rendering capabilities.

The framework offers three conceptual advantages. First, coordinates and momenta separate configuration-like information from motion-related dynamics. Second, an energy-based transition combines Hamiltonian evolution with learned dissipation. Third, multi-step latent consistency aligns the structured rollout with backbone trajectories over horizons where physical drift emerges. Together, these components form a plug-and-play transition module compatible with existing observation interfaces.

We evaluate \method{} in simulation and three embodied world-model settings. In simulation, it improves speed, acceleration, and jerk prediction while better preserving energy relative to an unstructured transition. On Ctrl-World for single-arm multi-view manipulation and EVAC for dual-arm single-view manipulation, it improves generation and logic metrics, supporting better visual-temporal and physical consistency. On Cosmos-Predict2.5-2B for single-arm single-view task execution, it increases success rates on LIBERO and Robocasa, demonstrating benefits for both prediction quality and downstream execution.

Our contributions are summarized as follows:

\begin{enumerate}[
    leftmargin=*,
    topsep=2pt,
    itemsep=1pt,
    parsep=0pt,
    partopsep=0pt
]
\item We propose a transition-centric view of world modeling: physical consistency should be built into the latent transition law, rather than added through auxiliary losses or post-hoc evaluation.
\item We introduce \method{}, a plug-and-play Hamiltonian latent transition framework that maps backbone latents into phase-space variables and evolves them with dissipative Hamiltonian dynamics for improved long-horizon stability.
\item We demonstrate consistent benefits across simulation and three embodied world-model settings, including physical-state prediction, better visual and logic metrics on Ctrl-World and EVAC, and higher downstream task success on LIBERO and Robocasa with Cosmos. 
\end{enumerate}
\vspace{-0.3cm}
\section{Related Work}
\vspace{-0.2cm}

\textbf{Generative and embodied video world models.}
Recent work has scaled world models into high-capacity video generators and interactive simulators. GAIA-1 and Genie demonstrated that generative models can support controllable, action-conditioned future prediction at scale \citep{hu2023gaia, bruce2024genie}. In robotics, manipulation-oriented models such as Multi-View Masked World Models, RoboDreamer, IRASim, Ctrl-World, and Cosmos further improved view consistency, controllability, and open-loop rollout quality for embodied interaction \citep{seo2023multiviewmaskedworldmodels, zhou2024robodreamerlearningcompositionalworld, zhu2025irasimfinegrainedworldmodel, guo2026ctrlworldcontrollablegenerativeworld, nvidia2025cosmosworldfoundationmodel}. Broader embodied-data and system efforts, such as AgiBot World, likewise reflect the field's shift toward scalable, manipulation-centered world modeling \citep{bu2025agibotworld}. Our work is complementary to this line: rather than proposing a new rendering or tokenization backbone, \method{} replaces the generic latent transition with a physically structured alternative. This distinction matters because current systems can achieve strong perceptual realism while still learning dynamics mostly implicitly, which leaves room for physical drift over long rollouts.

\textbf{Physics-structured neural dynamics and world models.}
Another line of work introduces explicit physical structure into learned dynamics. Hamiltonian Neural Networks, Hamiltonian Generative Networks, and Symplectic ODE-Nets showed that structure-preserving parameterizations can improve stability and extrapolation in dynamical systems \citep{greydanus2019hamiltonian, toth2020hamiltonian, zhong2020symplectic}. More recent world-model variants, including HaM-World, PH-Dreamer, and LaWM, inject Hamiltonian, port-Hamiltonian, or least-action principles into latent transitions \citep{tang2026hamworld, luan2026phdreamer, xiao2026lawm}. \method{} is most closely related to this family, but differs in emphasis: we aim for a drop-in transition design for embodied video world models that jointly captures conservative evolution, control, dissipation, and residual contact-rich effects, instead of focusing only on planner-oriented control benchmarks or on a single physical formalism.

\section{Preliminaries}

We briefly review two building blocks of our approach: learned world models for robotic systems, and Hamiltonian dynamics as an inductive bias for structured state transitions.

\subsection{World Models for Embodied Agents}
\label{subsec:wm_definition}
We learn a forward model in a low-dimensional latent space rather than directly from raw observations. An encoder $e_\phi$ maps an observation $o_t$ to a latent state $z_t$, a dynamics predictor $f_\theta$ advances $z_t$ to $z_{t+1}$ given action $a_t$, and a decoder $d_\psi$ reconstructs or predicts future observations $\hat{o}_t$. The model is trained by minimizing the multi-step prediction error, e.g.,
\begin{equation}
\mathcal{L} = \mathbb{E}_{(o_t,a_t)\sim\mathcal{D}} \sum_{t=1}^{T} \| o_t - d_\psi(z_t) \|^2 + \lambda \cdot \mathcal{R}(z_t),
\label{eq:wm_loss}
\end{equation}
where $\mathcal{R}(z_t)$ is a latent regularizer (e.g., KL divergence to a prior in variational settings).

Standard black-box dynamics trained this way tend to accumulate error and violate physical invariants over long rollouts, especially in contact-rich manipulation. To address this, we constrain $f_\theta$ to follow a controlled dissipative Hamiltonian structure in latent space, where $z_t$ is split into generalized coordinates $q_t$ and momenta $p_t$, and the transition is governed by a learnable energy function $H_\theta(q,p)$. This inductive bias enforces energy conservation (up to explicit dissipation and actuation) and symplectic geometry, which we formalize next.

\subsection{Hamiltonian Dynamics}
\label{subsec:hamiltonian_prelim}
To formalize the physical structure used by \method{}, we review Hamiltonian dynamics in a form suitable for latent-space world modeling. Unlike classical settings where $q, p$ are physical coordinates, in our framework they are \emph{learned latent phase variables} inferred from raw observations. Let
\begin{equation}
z_t = \begin{bmatrix} q_t \\ p_t \end{bmatrix} \in \mathbb{R}^{2d}
\end{equation}
denote a latent phase state, where $q_t \in \mathbb{R}^{d}$ represents generalized latent coordinates and $p_t \in \mathbb{R}^{d}$ represents generalized latent momenta. A Hamiltonian $H_\theta(q,p)$ (parameterized by a neural network) defines the total energy of the latent system. In the separable case, it is often written as
\begin{equation}
H_\theta(q,p) = T_\theta(p) + V_\theta(q),
\end{equation}
where $T_\theta(p)$ is the learned kinetic energy and $V_\theta(q)$ is the learned potential energy.

For a conservative latent system, the dynamics are governed by the canonical Hamilton equations
\begin{equation}
\dot q = \frac{\partial H_\theta}{\partial p}, \qquad
\dot p = -\frac{\partial H_\theta}{\partial q}.
\label{eq:canonical_hamilton_latent}
\end{equation}
Equivalently, by defining the symplectic matrix
\begin{equation}
J =
\begin{bmatrix}
0 & I \\
-I & 0
\end{bmatrix},
\end{equation}
the latent dynamics can be written compactly as
\begin{equation}
\dot z = J \nabla_z H_\theta(z),
\label{eq:hamiltonian_matrix_latent}
\end{equation}
where $\nabla_z H_\theta(z)$ denotes the energy gradient in latent phase space. Since $J^\top = -J$, Eq.~\eqref{eq:hamiltonian_matrix_latent} preserves the Hamiltonian along trajectories of an autonomous conservative latent system:
\begin{equation}
\frac{d}{dt} H_\theta(z_t)
= \nabla_z H_\theta(z_t)^\top \dot z_t
= \nabla_z H_\theta(z_t)^\top J \nabla_z H_\theta(z_t)
= 0.
\label{eq:conservative_energy_latent}
\end{equation}

Embodied manipulation, however, is not purely conservative: robot actions inject energy, while friction, damping, and contact dissipation remove it. To model this setting, we consider a \emph{controlled dissipative Hamiltonian} system in the latent space:
\begin{equation}
\dot z
= \bigl(J - R_\phi(z)\bigr)\nabla_z H_\theta(z) + B_\psi(z)u + r_\omega(z,u),
\label{eq:controlled_dissipative_hamiltonian_latent}
\end{equation}
where $R_\phi(z) \succeq 0$ is a learned dissipation matrix, $u$ is the robot control input, $B_\psi(z)$ maps controls into latent phase-space forces, and $r_\omega(z,u)$ captures residual non-conservative effects such as contact impulses, deformation, or modeling error. The corresponding energy change in the latent space is
\begin{equation}
\frac{d}{dt} H_\theta(z_t)
= - \nabla_z H_\theta(z_t)^\top R_\phi(z_t)\nabla_z H_\theta(z_t)
+ \nabla_z H_\theta(z_t)^\top B_\psi(z_t)u_t
+ \nabla_z H_\theta(z_t)^\top r_\omega(z_t,u_t).
\label{eq:controlled_energy_balance_latent}
\end{equation}
Equation~\eqref{eq:controlled_energy_balance_latent} makes explicit how conservative transport, dissipation, actuation, and residual contact effects contribute differently to the evolution of the latent energy, providing a strong inductive bias for long-horizon prediction.

For long-horizon prediction, the discretization of Eq.~\eqref{eq:controlled_dissipative_hamiltonian_latent} is as important as the continuous-time dynamics. A naive explicit Euler update can accumulate substantial geometric drift, whereas structure-preserving integrators better retain the phase-space organization of the underlying system \citep{zhong2020symplectic}. A standard symplectic Euler discretization of the conservative core is
\begin{equation}
p_{t+1} = p_t - \Delta t\, \nabla_q H_\theta(q_t, p_{t+1}), \qquad
q_{t+1} = q_t + \Delta t\, \nabla_p H_\theta(q_t, p_{t+1}),
\label{eq:symplectic_euler_latent}
\end{equation}
which can be extended with control, dissipation, and residual terms. In \method{}, observations are mapped to latent phase variables $(q_t, p_t)$ via $e_\phi$, the dominant transition is evolved by the controlled dissipative Hamiltonian rule, and additional residual channels $r_\omega$ absorb non-ideal effects that are unavoidable in real manipulation.

\section{Hamiltonian Latent World Models for Embodied Agents}
\label{sec:method}
\vspace{-.3cm}

We present \method{}, a Hamiltonian latent world model for embodied agents. Our framework follows the standard world-model architecture: an encoder maps observations to latent states, a structured dynamics module advances the latent state over time, and a decoder renders predictions back to observation space. The key departure from prior work lies in the dynamics module: instead of relying solely on unconstrained neural latent transitions, we introduce a Hamiltonian phase-space evolution that enforces physical inductive biases in the latent space.

\vspace{-.1cm}
\subsection{Why Hamiltonian Dynamics?}

World models learn latent dynamics that approximate the evolution of observed physical systems. A variational perspective provides a structured way to describe such dynamics through stationary paths of an action functional
$
\mathcal{S}[q] = \int_{t_0}^{t_1} L(q,\dot q;t)\,dt,
$
where $L$ is a Lagrangian. These paths satisfy the Euler--Lagrange equations
$
\frac{d}{dt}\frac{\partial L}{\partial \dot q}
=
\frac{\partial L}{\partial q}.
$

Hamiltonian mechanics reformulates this principle in phase space. Introducing generalized momenta $p=\partial L/\partial\dot q$, an invertible Legendre transform yields the Hamiltonian
$
H(q,p;t)=p^\top\dot q-L(q,\dot q;t)
$
and Hamilton's equations
$
\dot q=\frac{\partial H}{\partial p},
\quad
\dot p=-\frac{\partial H}{\partial q}.
$
This reformulation preserves the variational dynamics while making their symplectic structure explicit.

For latent world models, this connection motivates Hamiltonian dynamics as a structured inductive bias rather than a necessary consequence of prediction-error minimization. A learned Hamiltonian $H_\theta(q,p)$ constrains latent evolution through phase-space geometry while allowing the energy landscape to be learned from data. For robotic manipulation, we augment this conservative core with action-dependent forcing and learned dissipation $D_\theta(p)$ to account for external inputs, friction, and damping. The resulting controlled dissipative dynamics retain a symplectic conservative component while accommodating non-conservative effects.

\subsection{Latent Phase-Space Encoding}

\begin{figure}[t!]
  \setlength{\abovecaptionskip}{-0.1cm}
  \setlength{\belowcaptionskip}{-0.2cm}
  \vspace{-.1cm}
  \centering

    \includegraphics[width=0.85\columnwidth]{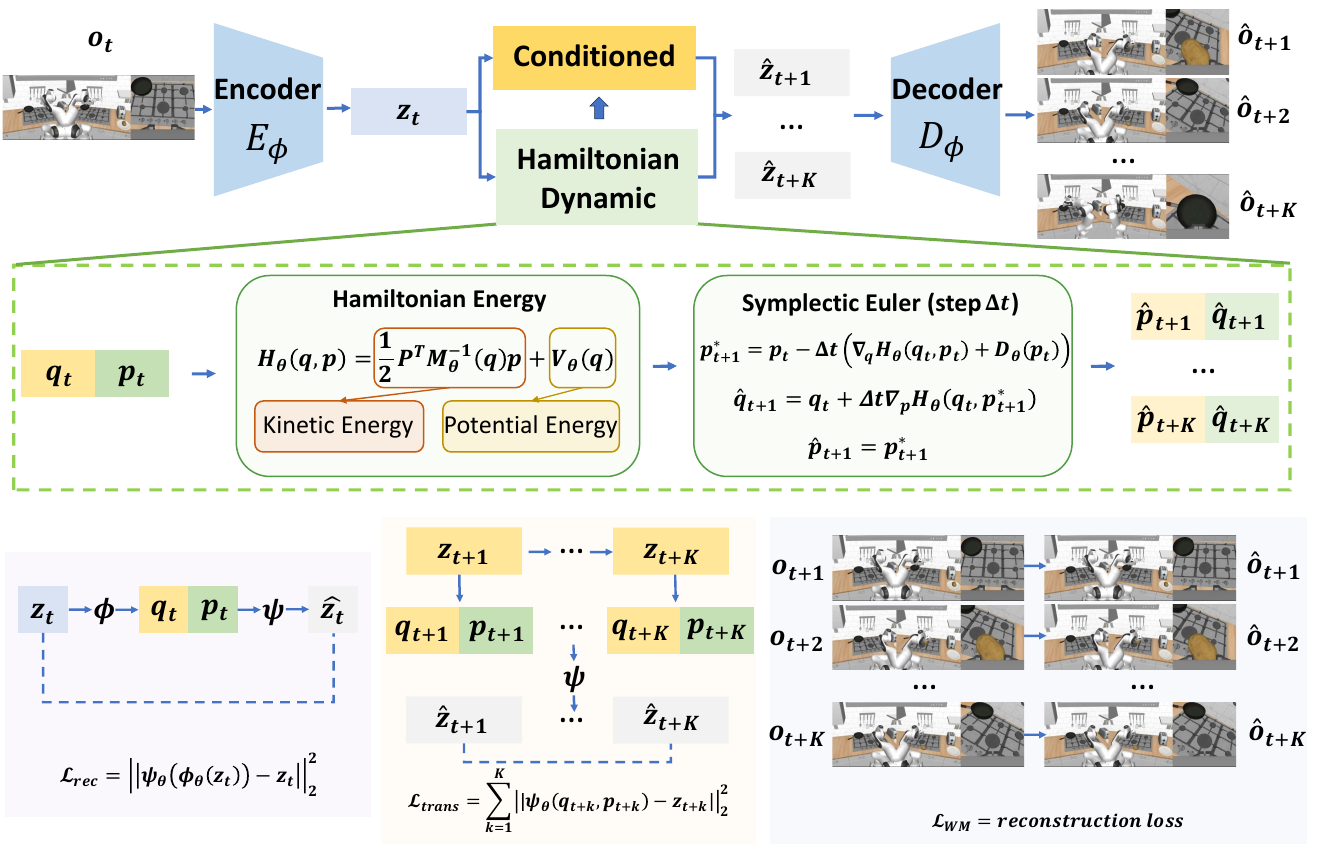}
  \caption{Overview of the proposed Hamiltonian latent dynamics framework. Given an observation $o_t$, the encoder $E_\phi$ maps it to a latent state $z_t$, which is decomposed into generalized coordinates and momenta, $(q_t,p_t)$, and evolved over multiple future steps using a learned hamiltonian energy function and a Symplectic Euler integrator. The resulting latent trajectory is mapped to latent representations and decoded by $D_\phi$ to predict future observations. Training is jointly guided by the latent reconstruction loss, latent transition loss, and world-model reconstruction loss.}
  \label{fig:method}
\end{figure}

As illustrated in Fig.~\ref{fig:method}, the backbone encoder $E_\phi$ first maps the observation $o_t$ to a latent representation
\begin{equation}
z_t = E_\phi(o_t),
\qquad
z_t \in \mathbb{R}^{d_z}.
\end{equation}
To introduce a structured dynamical representation, \method{} further maps the latent state $z_t$ into a Hamiltonian phase space consisting of generalized coordinates and momenta:
$
(q_t,p_t) = \phi_\theta(z_t),
\label{eq:phase_mapping}
$
where $q_t,p_t \in \mathbb{R}^{d}$ denote latent generalized coordinates and momenta, respectively.

This phase-space representation serves two purposes. First, it explicitly introduces the coordinate--momentum structure required by Hamiltonian mechanics, where physical dynamics are represented as a first-order system in $(q,p)$. Second, the explicit momentum variable provides an independent representation of velocity- and inertia-related information, which is important for modeling systems with non-negligible momentum and temporal persistence.

To maintain compatibility with the original backbone latent space, we introduce a learnable inverse mapping
$
\hat z_t = \psi_\theta(q_t,p_t),
\label{eq:phase_inverse}
$
which maps a phase-space state back to the latent representation used by the backbone decoder. Therefore, $\phi_\theta$ and $\psi_\theta$ form a bidirectional interface between the original latent space and the structured Hamiltonian phase space.

We parameterize the latent Hamiltonian using a separable kinetic--potential form:
\begin{equation}
H_\theta(q_t,p_t)
=
\frac{1}{2}
p_t^\top M_\theta^{-1}(q_t)p_t
+
V_\theta(q_t),
\label{eq:hamiltonian}
\end{equation}
where $M_\theta^{-1}(q_t)$ denotes a learned inverse mass matrix and $V_\theta(q_t)$ denotes the learned potential energy. The first term corresponds to kinetic energy, while the second captures potential energy.

The separable structure $T(q,p)+V(q)$ provides a physically motivated inductive bias while retaining the representational flexibility of neural networks. In particular, the quadratic dependence on momentum makes the momentum contribution to the kinetic energy explicit, while the configuration-dependent mass matrix $M_\theta(q_t)$ allows the effective inertia of the latent system to vary with its configuration. This is particularly suitable for articulated or highly nonlinear systems whose inertial properties depend on their state.

\vspace{.9cm}

\subsection{Hamiltonian Dynamics in Latent Space}

Given the Hamiltonian in Eq.~\eqref{eq:hamiltonian}, the latent dynamics are governed by a dissipative Hamiltonian system:
\begin{align}
\dot q_t
&=
\nabla_p H_\theta(q_t,p_t),
\label{eq:dyn_q}
\\
\dot p_t
&=
-\nabla_q H_\theta(q_t,p_t)
-
D_\theta(p_t),
\label{eq:dyn_p}
\end{align}
where $D_\theta(p_t)$ denotes a learned damping term.

The first terms in Eqs.~\eqref{eq:dyn_q}--\eqref{eq:dyn_p} describe the conservative Hamiltonian dynamics, while $D_\theta(p_t)$ accounts for dissipative effects such as friction and unmodeled energy loss. Following the formulation illustrated in Fig.~\ref{fig:method}, we use a momentum-dependent damping function of the form
$
D_\theta(p_t)
=
\gamma_\theta(p_t)\,p_t,
\label{eq:damping}
$
where $\gamma_\theta(\cdot)$ is a learned scalar-valued function.

This choice provides a simple and effective dissipation mechanism while preserving the Hamiltonian structure of the conservative component. Compared with a fully parameterized dissipation matrix, the momentum-proportional formulation introduces fewer additional degrees of freedom and reduces the complexity of learning, while remaining expressive enough to capture the dominant energy-loss effects in the target dynamics.

\subsection{Structure-Preserving Rollout}

To predict future latent states, we discretize the continuous dynamics using the semi-implicit symplectic Euler scheme shown in Fig.~\ref{fig:method}. Given the current phase-space state $(q_t,p_t)$, one rollout step is defined as
\begin{align}
p_{t+1}^{*}
&=
p_t
-
\Delta t
\left(
\nabla_q H_\theta(q_t,p_t)
+
D_\theta(p_t)
\right),
\label{eq:symplectic_p}
\\
\hat q_{t+1}
&=
q_t
+
\Delta t\,
\nabla_p H_\theta(q_t,p_{t+1}^{*}),
\label{eq:symplectic_q}
\\
\hat p_{t+1}
&=
p_{t+1}^{*}.
\label{eq:symplectic_p_hat}
\end{align}

The updated phase-space state is therefore
$
(\hat q_{t+1},\hat p_{t+1}),
$
which is recursively used to generate a $K$-step prediction trajectory:
$
\left\{
(\hat q_{t+k},\hat p_{t+k})
\right\}_{k=1}^{K}.
\label{eq:phase_rollout}
$

The key feature of the semi-implicit symplectic Euler scheme is that the position update evaluates $\nabla_p H_\theta$ using the newly updated momentum $p_{t+1}^{*}$. Compared with a naive explicit Euler discretization, this update better preserves the geometric structure of the conservative Hamiltonian system and reduces numerical energy drift over long-horizon rollouts. The damping term is evaluated explicitly, which provides a practical compromise between numerical structure preservation and computational efficiency.

After the phase-space rollout, each predicted state is projected back into the original backbone latent space through the inverse phase-space mapping:
\begin{equation}
\hat z_{t+k}
=
\psi_\theta
\left(
\hat q_{t+k},
\hat p_{t+k}
\right),
\qquad
k=1,\ldots,K.
\label{eq:latent_decode}
\end{equation}

The corresponding future observations are then obtained using the backbone decoder:
\begin{equation}
\hat o_{t+k}
=
D_\phi(\hat z_{t+k}),
\qquad
k=1,\ldots,K.
\label{eq:obs_decode}
\end{equation}

Thus, as illustrated in Fig.~\ref{fig:method}, the complete prediction process consists of three stages:
(i) encoding the current observation into the backbone latent space,
(ii) evolving the latent state in the structured Hamiltonian phase space, and
(iii) mapping the predicted phase states back to the latent and observation spaces.

\subsection{Training Objective}

The Hamiltonian dynamics module is trained jointly with the backbone world model using the following multi-objective loss:
\begin{equation}
\mathcal{L}_{\mathrm{total}}
=
\mathcal{L}_{\mathrm{WM}}
+
\lambda_{\mathrm{rec}}
\mathcal{L}_{\mathrm{rec}}
+
\lambda_{\mathrm{trans}}
\mathcal{L}_{\mathrm{trans}},
\label{eq:total_loss}
\end{equation}
where $\mathcal{L}*{\mathrm{WM}}$ denotes the observation-level reconstruction loss, $\mathcal{L}*{\mathrm{rec}}$ enforces consistency between the latent and phase-space representations, and $\mathcal{L}_{\mathrm{trans}}$ supervises the multi-step Hamiltonian rollout.

First, the phase-space reconstruction loss is defined as
$
\mathcal{L}_{\mathrm{rec}}
=
\left\|
\psi_\theta
\left(
\phi_\theta(z_t)
\right)
-
z_t
\right\|_2^2.
\label{eq:rec_loss}
$
This objective encourages $\phi_\theta$ and $\psi_\theta$ to behave as approximately inverse mappings. It ensures that the learned phase-space representation remains compatible with the backbone latent space and prevents the model from learning a dynamically convenient but semantically disconnected latent coordinate system.

Second, we introduce a transition consistency loss over the entire prediction horizon:
\begin{equation}
\mathcal{L}_{\mathrm{trans}}
=
\sum_{k=1}^{K}
\left\|
\psi_\theta
\left(
\hat q_{t+k},
\hat p_{t+k}
\right)
-
z_{t+k}
\right\|_2^2,
\label{eq:trans_loss}
\end{equation}
where the target future latent state is obtained from the ground-truth observation:
\begin{equation}
z_{t+k}
=
E_\phi(o_{t+k}),
\qquad
k=1,\ldots,K.
\label{eq:gt_future_latent}
\end{equation}

The transition consistency loss directly constrains the Hamiltonian rollout to remain on the latent manifold induced by the backbone encoder. In contrast to one-step supervision, the multi-step formulation penalizes accumulated prediction error and discourages the dynamics model from drifting into phase-space regions that are dynamically plausible under the learned Hamiltonian but inconsistent with the actual latent trajectories.

Finally, the observation-level world-model loss is computed between the predicted and ground-truth future observations:
$
\mathcal{L}_{\mathrm{WM}}
=
\sum_{k=1}^{K}
\ell_{\mathrm{WM}}
\left(
\hat o_{t+k},
o_{t+k}
\right),
\label{eq:wm_reconstruction_loss}
$
where $\ell_{\mathrm{WM}}$ denotes the reconstruction loss adopted by the backbone world model.

Together, these three losses establish consistency across the phase-space, latent-space, and observation-space levels. Specifically, $\mathcal{L}_{\mathrm{rec}}$ constrains the phase-space representation, $\mathcal{L}_{\mathrm{trans}}$ constrains the temporal evolution of the structured latent dynamics, and $\mathcal{L}_{\mathrm{WM}}$ ensures that the resulting predictions remain accurate in observation space. All components are differentiable and optimized jointly through backpropagation through the multi-step Hamiltonian rollout.

\section{Experiments}
\vspace{-.2cm}

We evaluate \method{} from three complementary perspectives: physical-state prediction and energy consistency in simulation, visual generation quality, and long-horizon physical consistency and downstream task success in real-world rollouts. In simulation, we test whether \method{} captures underlying physical regularities beyond short-range correlations. For real-world evaluation, we consider three world-model families: COSMOS for single-arm single-view prediction \citep{nvidia2025cosmosworldfoundationmodel}, Ctrl-World for single-arm multi-view manipulation \citep{guo2026ctrlworldcontrollablegenerativeworld}, and EVAC \citep{jiang2025enerverse} for dual-arm single-view manipulation, using each backbone's standard visual generation metrics. We emphasize Ctrl-World for long-horizon analysis, as multi-view observations reveal temporal and geometric inconsistencies. Across experiments, we preserve the original observation interface and other backbone components, replacing only the latent transition with \method{} where applicable and comparing primarily against the corresponding vanilla backbone.

\subsection{Simulation Experiments}
\label{settings}
\vspace{-.1cm}

To isolate the effect of latent transition modeling from high-dimensional visual rendering, we first evaluate \method{} in simulation using SimVP as the baseline, where accurate physical annotations enable direct assessment of learned dynamics. Detailed simulation settings are provided in Appendix~\ref{simulation}.

\begin{figure}[t]
    \centering
\includegraphics[width=\linewidth]{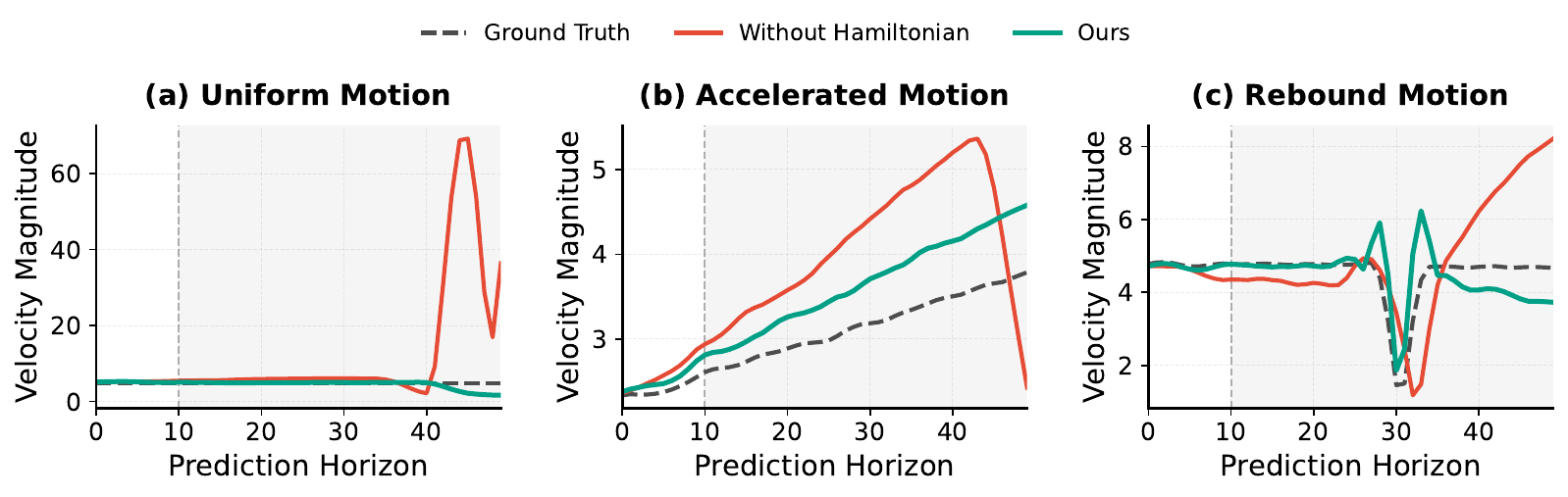}
\vspace{-0.6cm}
    \caption{Qualitative comparison of the predicted velocity curves.}
    \vspace{-0.3cm}
    \label{fig:velocity_curve}
\end{figure}

\textbf{Instantaneous Physical State Prediction.} We evaluate whether \method{} better captures instantaneous physical states, including position, speed, acceleration, and jerk. As shown in Table~\ref{tab:overall_motion_results}, \method{} consistently reduces prediction errors across all metrics and rollout horizons. The predicted velocity trajectories in Fig.~\ref{fig:velocity_curve} also remain closer to the ground truth, indicating that Hamiltonian structuring improves the accuracy and stability of the learned dynamics.

\vspace{-.4cm}
\begin{table}[htbp]
\centering
\caption{Instantaneous physical-state prediction under different rollout horizons. 
All models are trained with a 10-step prediction horizon. Lower values indicate better performance.}
\label{tab:overall_motion_results}
{
\begin{tabular}{cccccc}
\toprule
Method & Step
& Pos. RMSE
& Vel. RMSE
& Acc. RMSE
& Jerk RMSE \\
\midrule

Vanilla & 30
& 6.217
& 3.455
& 10.966
& 14.135 \\

\rowcolor{blue!8}
\method{} & 30
& \textbf{6.212}
& \textbf{3.372}
& \textbf{9.235}
& \textbf{11.401} \\

Vanilla & 50
& 5.590
& 2.545
& 8.332
& 12.329 \\

\rowcolor{blue!8}
\method{} & 50
& \textbf{5.524}
& \textbf{2.525}
& \textbf{7.483}
& \textbf{10.750} \\

\bottomrule
\end{tabular}
}
\end{table}

\begin{figure}[t]
    \centering
    \includegraphics[width=0.85\linewidth]{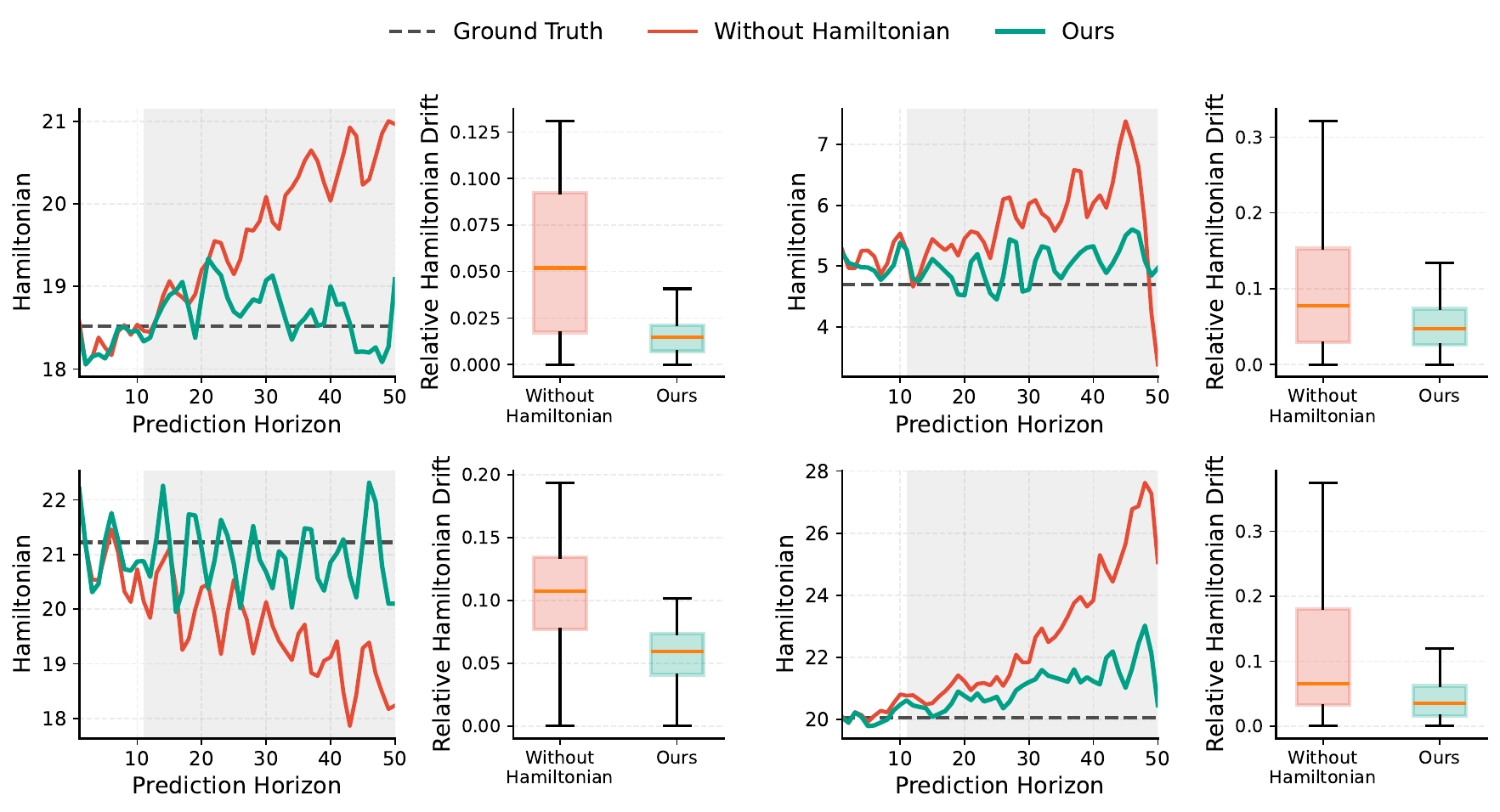}
    \vspace{-.4cm}
    \caption{
    Comparison of energy conservation across different methods.
    }
    \label{fig:energy_curve}
\end{figure}

\textbf{Energy Conservation.} We further evaluate long-horizon physical consistency through Hamiltonian energy drift. As shown in Fig.~\ref{fig:energy_curve}, \method{} maintains energy trajectories substantially closer to the ground truth, while the non-Hamiltonian variant exhibits larger drift and oscillations. This demonstrates that the proposed structured dynamics better preserve global physical consistency over long rollouts.
\vspace{-.4cm}

\begin{table}[h!]
\centering
\caption{Quantitative comparison between \method{} and Vanilla on Ctrl-World.}
\begin{tabular}{c|ccc|ccc}
\hline
Method & nDTW & PSNR & BLEU & Scene (\%) & CLIP (\%) & Logic (\%)\\
\hline
Vanilla & 0.670 & 10.13 & 0.767 & 90.88 & 78.31 & 93.3 \\

\rowcolor{blue!8}
\method{} & \textbf{0.735} & \textbf{10.16} & \textbf{0.802}
& \textbf{90.96} & \textbf{78.96} & \textbf{96.7} \\

\hline
\end{tabular}
\label{tab:ctrl_world_hami}
\end{table}

\begin{table}[htbp]
\centering
\caption{Comparison between \method{} and Vanilla models. Metrics are collected from two evaluation settings on EVAC.}
\begin{tabular}{c|ccc|ccc}
\hline
Method & nDTW & PSNR & BLEU & Scene (\%) & CLIP (\%) & Logic (\%)\\
\hline
Vanilla & 0.282 & 16.61 & 0.104 & 91.3 & 77.5 & 80.0 \\

\rowcolor{blue!8}
\method{} & \textbf{0.283} & \textbf{16.64} & \textbf{0.111}
& \textbf{92.2} & \textbf{78.4} & \textbf{86.7} \\

\hline
\end{tabular}
\label{tab:hami_vs_vanilla}
\end{table}

\vspace{-.5cm}
\begin{table}[h!]
\centering
\caption{Overall success rate comparison on LIBERO and Robocasa benchmarks.}
\begin{tabular}{c|ccccc|c}
\hline
\multirow{2}{*}{Method}
& \multicolumn{5}{c|}{LIBERO}
& \multicolumn{1}{c}{Robocasa} \\
\cline{2-7}
& Spatial (\%) & Object (\%) & Goal (\%) & Long (\%) & Avg (\%) & Avg (\%)\\
\hline
Vanilla & 97.6 & 99.8 & 96.2 & 97.6 & 97.8 & 66.7 \\

\rowcolor{blue!8}
\method{} & \textbf{98.6} & \textbf{99.8} & \textbf{98.0}
& \textbf{98.2} & \textbf{98.7} & \textbf{68.5} \\

\hline
\end{tabular}
\label{tab:overall_libero_robocasa}
\end{table}

\subsection{Real-World Large World-Model Experiments: Visual Generation Quality}

We evaluate \method{} as a drop-in transition module in Ctrl-World~\citep{guo2026ctrlworldcontrollablegenerativeworld} and EVAC~\citep{jiang2025enerverse}.
As shown in Table~\ref{tab:ctrl_world_hami}, nDTW increases from 0.670 to 0.735, BLEU from 0.767 to 0.802, and Logic from 93.3\% to 96.7\% on Ctrl-World.
On EVAC, Table~\ref{tab:hami_vs_vanilla} shows that BLEU increases from 0.104 to 0.111 and Logic from 80.0\% to 86.7\%, with improvements in the remaining reported metrics.
These results support improved multi-view visual consistency and generation coherence for bimanual interactions across different architectures and embodiment settings.
Detailed settings and evaluation protocols appear in Appendix~\ref{app:real_world_visual}.

\vspace{-.2cm}
\subsection{Real-World Large World-Model Experiments: Task Execution Success Rate}

We integrate \method{} into \textit{Cosmos-Predict2.5-2B} and evaluate downstream task execution on LIBERO~\citep{liu2023libero} and RoboCasa~\citep{nasiriany2024robocasa}. As shown in Table~\ref{tab:overall_libero_robocasa}, average success increases from 97.8\% to 98.7\% on LIBERO, with matching or improved performance across all four suites, and from 66.7\% to 68.5\% on RoboCasa. These results support the usefulness of Hamiltonian transitions for downstream embodied execution. Evaluation protocols, task-level results across seeds, and qualitative examples appear in Appendix~\ref{app:cosmos_execution}.

\section{Conclusion}
\vspace{-.2cm}
We presented \method{}, a Hamiltonian transition framework that combines controlled dissipative dynamics with a residual branch to improve long-horizon physical consistency in embodied world models. Experiments across Ctrl-World, COSMOS, and EVAC demonstrate benefits in prediction quality, physical consistency, and downstream task execution. Several limitations remain. Due to limited computational resources, we could not evaluate more backbones or conduct broader experiments across tasks, training scales, and random seeds. Moreover, learned phase variables need not correspond to identifiable physical quantities, so Hamiltonian structure alone does not guarantee physically correct predictions. Dissipation, external control, and the residual branch also limit the conservation properties of the complete dynamics, particularly during contact-rich interactions. Finally, the current evaluations do not establish robustness to substantially different environments or rollout horizons beyond those tested. Broader evaluation and stronger characterization of the learned dynamics are therefore needed.

\section*{AI Use Statement}

We used large language models (LLMs) to assist with language polishing,
including improving grammar, readability, and presentation of the manuscript.
LLMs were not used to generate experimental results or make scientific
conclusions. All technical content, mathematical formulations, experimental
designs, analyses, and reported results were developed and verified by the
authors. The authors take full responsibility for the content of the paper.

\section*{Reproducibility Statement}

We provide an anonymized code repository containing the implementation of
the proposed Hamiltonian World Model (HWM) to support reproducibility.
The paper and appendix describe the model architecture, latent phase-space
parameterization, Hamiltonian dynamics, numerical integration scheme,
training objectives, experimental settings, evaluation protocols, and
benchmark configurations used in our experiments.

We report the relevant training and evaluation horizons, baseline
configurations, benchmark protocols, and random-seed evaluations where
applicable. Our experiments cover both controlled simulation and multiple
embodied world-model backbones, with the corresponding experimental details
provided in the appendix. Together with the anonymized implementation, these
details are intended to enable reproduction of the main results reported in
the paper.

\section*{Ethics Statement}
This work does not involve human subjects, sensitive personal data, or deployment on physical robots that could directly endanger users. Our experiments focus on improving the reliability of learned world models for embodied interaction. More physically consistent world models could help reduce unsafe action proposals by making model-based evaluation more faithful, but they could also strengthen autonomous systems in ways that require responsible downstream use. We therefore emphasize transparent evaluation, careful reporting, and reproducible methodology.

To promote reproducibility and transparency, we provide detailed experimental settings, algorithmic descriptions, and anonymized code. Our contribution is methodological and aims to improve the reliability of embodied world modeling rather than bypass safety constraints.

\bibliography{iclr2026_conference}

@article{nasiriany2024robocasa,
  title={Robocasa: Large-scale simulation of everyday tasks for generalist robots},
  author={Nasiriany, Soroush and Maddukuri, Abhiram and Zhang, Lance and Parikh, Adeet and Lo, Aaron and Joshi, Abhishek and Mandlekar, Ajay and Zhu, Yuke},
  journal={arXiv preprint arXiv:2406.02523},
  year={2024}
}

@article{kim2026cosmos,
  title={Cosmos policy: Fine-tuning video models for visuomotor control and planning},
  author={Kim, Moo Jin and Gao, Yihuai and Lin, Tsung-Yi and Lin, Yen-Chen and Ge, Yunhao and Lam, Grace and Liang, Percy and Song, Shuran and Liu, Ming-Yu and Finn, Chelsea and others},
  journal={arXiv preprint arXiv:2601.16163},
  year={2026}
}

@article{liu2023libero,
  title={Libero: Benchmarking knowledge transfer for lifelong robot learning},
  author={Liu, Bo and Zhu, Yifeng and Gao, Chongkai and Feng, Yihao and Liu, Qiang and Zhu, Yuke and Stone, Peter},
  journal={Advances in Neural Information Processing Systems},
  volume={36},
  pages={44776--44791},
  year={2023}
}

@article{yue2025ewmbench,
  title={Ewmbench: Evaluating scene, motion, and semantic quality in embodied world models},
  author={Yue, Hu and Huang, Siyuan and Liao, Yue and Chen, Shengcong and Zhou, Pengfei and Chen, Liliang and Yao, Maoqing and Ren, Guanghui},
  journal={arXiv preprint arXiv:2505.09694},
  year={2025}
}

@article{jiang2025enerverse,
  title={Enerverse-ac: Envisioning embodied environments with action condition},
  author={Jiang, Yuxin and Chen, Shengcong and Huang, Siyuan and Chen, Liliang and Zhou, Pengfei and Liao, Yue and He, Xindong and Liu, Chiming and Li, Hongsheng and Yao, Maoqing and others},
  journal={arXiv preprint arXiv:2505.09723},
  year={2025}
}

@misc{tang2026hamworld,
  title={HaM-World: Soft-Hamiltonian World Models with Selective Memory for Planning},
  author={Tang, Haoyun and Cui, Haodong and Xu, Keyao and Wang, Kun and Mei, Zhandong},
  year={2026},
  eprint={2605.05951},
  archivePrefix={arXiv},
  primaryClass={cs.AI},
  url={https://arxiv.org/abs/2605.05951},
}

@misc{luan2026phdreamer,
  title={PH-Dreamer: A Physics-Driven World Model via Port-Hamiltonian Generative Dynamics},
  author={Luan, Xueyu and Shi, Chenwei},
  year={2026},
  eprint={2605.18303},
  archivePrefix={arXiv},
  primaryClass={cs.LG},
  url={https://arxiv.org/abs/2605.18303},
}

@misc{xiao2026lawm,
  title={LaWM: Least Action World Models for Long-Horizon Physical Consistency from Visual Observations},
  author={Xiao, Qixin and Ghaffari, Maani},
  year={2026},
  eprint={2605.08279},
  archivePrefix={arXiv},
  primaryClass={cs.LG},
  url={https://arxiv.org/abs/2605.08279},
}

@misc{jiang2026robowmbench,
  title={RoboWM-Bench: A Benchmark for Evaluating World Models in Robotic Manipulation},
  author={Jiang, Feng and Chen, Yang and Xu, Kyle and Liu, Yuchen and Wang, Haifeng and Shen, Zhenhao and Lu, Jasper and Huang, Shengze and Wang, Yuanfei and Xie, Chen and Wu, Ruihai},
  year={2026},
  eprint={2604.19092},
  archivePrefix={arXiv},
  primaryClass={cs.RO},
  url={https://arxiv.org/abs/2604.19092},
}

@misc{wang2026eva,
  title={EVA: Aligning Video World Models with Executable Robot Actions via Inverse Dynamics Rewards},
  author={Wang, Ruixiang and Liu, Qingming and Deng, Yueci and Liu, Guiliang and Liu, Zhen and Jia, Kui},
  year={2026},
  eprint={2603.17808},
  archivePrefix={arXiv},
  primaryClass={cs.RO},
  url={https://arxiv.org/abs/2603.17808},
}

@article{ha2018worldmodels,
  title={World Models},
  author={Ha, David and Schmidhuber, J{\"u}rgen},
  journal={arXiv preprint arXiv:1803.10122},
  year={2018}
}

@inproceedings{hafner2019planet,
  title={Learning Latent Dynamics for Planning from Pixels},
  author={Hafner, Danijar and Lillicrap, Timothy and Fischer, Ian and Villegas, Ruben and Ha, David and Lee, Honglak and Davidson, James},
  booktitle={Proceedings of the 36th International Conference on Machine Learning},
  pages={2555--2565},
  year={2019},
  organization={PMLR}
}

@inproceedings{hafner2020dreamer,
  title={Dream to Control: Learning Behaviors by Latent Imagination},
  author={Hafner, Danijar and Lillicrap, Timothy and Ba, Jimmy and Norouzi, Mohammad},
  booktitle={International Conference on Learning Representations},
  year={2020}
}

@article{hafner2023dreamerv3,
  title={Mastering Diverse Domains through World Models},
  author={Hafner, Danijar and Pasukonis, Jurgis and Ba, Jimmy and Lillicrap, Timothy},
  journal={arXiv preprint arXiv:2301.04104},
  year={2023}
}

@article{hu2023gaia,
  title={GAIA-1: A Generative World Model for Autonomous Driving},
  author={Hu, Anthony and Russell, Lloyd and Yeo, Hudson and Murez, Zak and Fedoseev, George and Kendall, Alex and Shotton, Jamie and Corrado, Gianluca},
  journal={arXiv preprint arXiv:2309.17080},
  year={2023}
}

@inproceedings{bruce2024genie,
  title={Genie: Generative Interactive Environments},
  author={Bruce, Jake and Dennis, Michael and Edwards, Ashley and Parker-Holder, Jack and Shi, Yuge and Hughes, Edward and Lai, Matthew and Mavalankar, Aditi and Steigerwald, Richie and Apps, Chris and Aytar, Yusuf and Bechtle, Sarah and Behbahani, Feryal and Chan, Stephanie and Heess, Nicolas and Gonzalez, Luis and Osindero, Simon and Ozair, Sherjil and Reed, Scott and Zhang, Jingwei and Zolna, Konrad and Clune, Jeff and de Freitas, Nando and Singh, Satinder and Rockt{\"a}schel, Tim},
  booktitle={Proceedings of the 41st International Conference on Machine Learning},
  year={2024}
}

@inproceedings{greydanus2019hamiltonian,
  title={Hamiltonian Neural Networks},
  author={Greydanus, Samuel and Dzamba, Misko and Yosinski, Jason},
  booktitle={Advances in Neural Information Processing Systems},
  volume={32},
  year={2019}
}

@inproceedings{toth2020hamiltonian,
  title={Hamiltonian Generative Networks},
  author={Toth, Peter and Rezende, Danilo J. and Jaegle, Andrew and Racani{\`e}re, S{\'e}bastien and Botev, Aleksandar and Higgins, Irina},
  booktitle={International Conference on Learning Representations},
  year={2020}
}

@inproceedings{zhong2020symplectic,
  title={Symplectic ODE-Net: Learning Hamiltonian Dynamics with Control},
  author={Zhong, Yaofeng Desmond and Dey, Biswadip and Chakraborty, Amit},
  booktitle={International Conference on Learning Representations},
  year={2020}
}

@misc{seo2023multiviewmaskedworldmodels,
  title={Multi-View Masked World Models for Visual Robotic Manipulation},
  author={Seo, Younggyo and Kim, Junsu and James, Stephen and Lee, Kimin and Shin, Jinwoo and Abbeel, Pieter},
  year={2023},
  eprint={2302.02408},
  archivePrefix={arXiv},
  primaryClass={cs.RO},
  url={https://arxiv.org/abs/2302.02408},
}

@misc{zhou2024robodreamerlearningcompositionalworld,
  title={RoboDreamer: Learning Compositional World Models for Robot Imagination},
  author={Zhou, Siyuan and Du, Yilun and Chen, Jiaben and Li, Yandong and Yeung, Dit-Yan and Gan, Chuang},
  year={2024},
  eprint={2404.12377},
  archivePrefix={arXiv},
  primaryClass={cs.RO},
  url={https://arxiv.org/abs/2404.12377},
}

@inproceedings{zhu2025irasimfinegrainedworldmodel,
  title={IRASim: A Fine-Grained World Model for Robot Manipulation},
  author={Zhu, Fangqi and Wu, Hongtao and Guo, Song and Liu, Yuxiao and Cheang, Chilam and Kong, Tao},
  booktitle={Proceedings of the IEEE/CVF International Conference on Computer Vision},
  year={2025},
  eprint={2406.14540},
  archivePrefix={arXiv},
  primaryClass={cs.RO},
  url={https://arxiv.org/abs/2406.14540},
}

@misc{nvidia2025cosmosworldfoundationmodel,
  title={Cosmos World Foundation Model Platform for Physical AI},
  author={{{NVIDIA}} and Niket Agarwal and Arslan Ali and Maciej Bala and Yogesh Balaji and Erik Barker and Tiffany Cai and Prithvijit Chattopadhyay and Yongxin Chen and Yin Cui and Yifan Ding and Daniel Dworakowski and Jiaojiao Fan and Michele Fenzi and Francesco Ferroni and Sanja Fidler and Dieter Fox and Songwei Ge and Yunhao Ge and Jinwei Gu and Siddharth Gururani and Ethan He and Jiahui Huang and Jacob Huffman and Pooya Jannaty and Jingyi Jin and Seung Wook Kim and Gergely Kl{\'a}r and Grace Lam and Shiyi Lan and Laura Leal-Taixe and Anqi Li and Zhaoshuo Li and Chen-Hsuan Lin and Tsung-Yi Lin and Huan Ling and Ming-Yu Liu and Xian Liu and Alice Luo and Qianli Ma and Hanzi Mao and Kaichun Mo and Arsalan Mousavian and Seungjun Nah and Sriharsha Niverty and David Page and Despoina Paschalidou and Zeeshan Patel and Lindsey Pavao and Morteza Ramezanali and Fitsum Reda and Xiaowei Ren and Vasanth Rao Naik Sabavat and Ed Schmerling and Stella Shi and Bartosz Stefaniak and Shitao Tang and Lyne Tchapmi and Przemek Tredak and Wei-Cheng Tseng and Jibin Varghese and Hao Wang and Haoxiang Wang and Heng Wang and Ting-Chun Wang and Fangyin Wei and Xinyue Wei and Jay Zhangjie Wu and Jiashu Xu and Wei Yang and Lin Yen-Chen and Xiaohui Zeng and Yu Zeng and Jing Zhang and Qinsheng Zhang and Yuxuan Zhang and Qingqing Zhao and Artur Zolkowski},
  year={2025},
  eprint={2501.03575},
  archivePrefix={arXiv},
  primaryClass={cs.CV},
  url={https://arxiv.org/abs/2501.03575},
}

@misc{guo2026ctrlworldcontrollablegenerativeworld,
  title={Ctrl-World: A Controllable Generative World Model for Robot Manipulation},
  author={Guo, Yanjiang and Shi, Lucy Xiaoyang and Chen, Jianyu and Finn, Chelsea},
  year={2025},
  eprint={2510.10125},
  archivePrefix={arXiv},
  primaryClass={cs.RO},
  url={https://arxiv.org/abs/2510.10125},
}

@article{sutton1991dyna,
  title={Dyna, an Integrated Architecture for Learning, Planning, and Reacting},
  author={Sutton, Richard S.},
  journal={ACM SIGART Bulletin},
  volume={2},
  number={4},
  pages={160--163},
  year={1991},
  publisher={ACM}
}

@misc{bu2025agibotworld,
  title={AgiBot World Colosseo: A Large-scale Manipulation Platform for Scalable and Intelligent Embodied Systems},
  author={{{AgiBot-World-Contributors}} and Qingwen Bu and Jisong Cai and Li Chen and Xiuqi Cui and Yan Ding and Siyuan Feng and Shenyuan Gao and Xindong He and Xuan Hu and Xu Huang and Shu Jiang and Yuxin Jiang and Cheng Jing and Hongyang Li and Jialu Li and Chiming Liu and Yi Liu and Yuxiang Lu and Jianlan Luo and Ping Luo and Yao Mu and Yuehan Niu and Yixuan Pan and Jiangmiao Pang and Yu Qiao and Guanghui Ren and Cheng Ruan and Jiaqi Shan and Yongjian Shen and Chengshi Shi and Mingkang Shi and Modi Shi and Chonghao Sima and Jianheng Song and Huijie Wang and Wenhao Wang and Dafeng Wei and Chengen Xie and Guo Xu and Junchi Yan and Cunbiao Yang and Lei Yang and Shukai Yang and Maoqing Yao and Jia Zeng and Chi Zhang and Qinglin Zhang and Bin Zhao and Chengyue Zhao and Jiaqi Zhao and Jianchao Zhu},
  year={2025},
  eprint={2503.06669},
  archivePrefix={arXiv},
  primaryClass={cs.RO},
  url={https://arxiv.org/abs/2503.06669},
}
\bibliographystyle{iclr2026_conference}

\newpage
\appendix

\section{Related Work}

\textbf{Latent world models for planning and control.}
World models have long served as learned simulators for sequential decision making, from Dyna and World Models to PlaNet, Dreamer, and DreamerV3 \citep{sutton1991dyna, ha2018worldmodels, hafner2019planet, hafner2020dreamer, hafner2023dreamerv3}. These works established that compact latent dynamics can support imagination-based planning and control, but they were primarily developed for reinforcement learning settings with comparatively simple observation and interaction structure. \method{} builds on this planner-facing latent-transition perspective, but targets embodied manipulation scenarios in which viewpoint changes, contact events, and action-conditioned rollouts place much stronger demands on long-horizon physical consistency.

\textbf{Embodiment-grounded evaluation of world models.}
As world models are increasingly used for robot learning, evaluation has moved beyond appearance metrics toward action-conditioned usefulness and executability. Recent studies argue that generated futures should support policy ranking or downstream control rather than merely look plausible \citep{guo2026ctrlworldcontrollablegenerativeworld, wang2026eva}. RoboWM-Bench makes this embodiment-grounded perspective explicit by evaluating whether generated manipulation videos can be translated into successful robot actions \citep{jiang2026robowmbench}. Our evaluation protocol follows this shift: we assess \method{} not only by predictive quality, but by whether its structured transitions improve long-horizon physical consistency and action usefulness across distinct embodied world-model families.

\section{Simulation Experiments}
\label{simulation}
\vspace{-.1cm}

\begin{table}[t]
    \centering
    \caption{Quantitative comparison of video prediction under different training and evaluation horizons. 
    Lower values indicate better performance.}
\label{tab:rollout_prediction}
    \begin{tabular}{c c cc cc}
        \toprule
        \multirow{2}{*}{Train Horizon} &
        \multirow{2}{*}{Test Horizon} &
        \multicolumn{2}{c}{MSE $\downarrow$} &
        \multicolumn{2}{c}{MAE $\downarrow$} \\
        \cmidrule(lr){3-4} \cmidrule(lr){5-6}
        & & Vanilla & \method{} & Vanilla & \method{} \\
        \midrule
        10 & 10 & 2.76  & \textbf{2.56}  & 23.88  & \textbf{21.27} \\
        10 & 30 & 25.38 & \textbf{23.82} & 53.56  & \textbf{49.27} \\
        10 & 50 & 55.57 & \textbf{51.18} & 87.59  & \textbf{80.37} \\
        10 & 60 & 62.32 & \textbf{53.03} & 95.32  & \textbf{82.58} \\
        \midrule
        30 & 30 & 25.81 & \textbf{24.38} & 79.36  & \textbf{73.31} \\
        30 & 50 & 37.55 & \textbf{33.65} & 94.54  & \textbf{87.84} \\
        30 & 60 & 44.59 & \textbf{38.93} & 102.55 & \textbf{94.60} \\
        \bottomrule
    \end{tabular}
\end{table}

To separate transition dynamics modeling from high-dimensional rendering complexity, we first conduct controlled experiments in a simulated environment. Because the simulator provides clean action-conditioned trajectories together with accurate physical annotations, it enables a more direct test of whether the learned latent dynamics preserve the underlying physical structure. Under this setting, we adopt SimVP as the baseline and investigate the effectiveness of incorporating the proposed Hamiltonian World Model (\method{}) . We evaluate \method{} from three complementary perspectives: long-horizon prediction quality, instantaneous physical-state prediction, and energy conservation.

\paragraph{Long-Horizon Prediction Quality.}
We first investigate whether incorporating Hamiltonian structure improves the overall quality of predicted trajectories. Specifically, we train the models with prediction horizons of $10$ and $30$ frames, and roll them out to different test horizons to evaluate both in-distribution prediction and long-horizon extrapolation. The quantitative results are reported in Table~\ref{tab:rollout_prediction}. Across all combinations of training and test horizons, \method{} consistently achieves lower MSE and MAE than the vanilla baseline, showing that imposing Hamiltonian structure also improves visual prediction quality. More importantly, the advantage becomes more pronounced under long-horizon rollout. For example, when trained with a $10$-frame prediction horizon and extrapolated to $60$ frames, \method{} reduces the MSE from $62.32$ to $53.03$ and the MAE from $95.32$ to $82.58$, corresponding to relative reductions of $14.9\%$ and $13.4\%$, respectively. These results suggest that Hamiltonian modeling mitigates the accumulation of dynamics errors during rollout, leading to more stable and accurate long-term predictions.

\paragraph{Instantaneous Physical State Prediction.}
Beyond overall prediction quality, we further evaluate the accuracy of instantaneous physical-state prediction, including object position, speed, acceleration, and jerk.  For each generated rollout, we first estimate the object center in every frame and compute velocity using finite differences between consecutive positions. Higher-order physical quantities, including acceleration and jerk, are then derived accordingly. 
Our evaluation includes the full \method{}, an ablated variant without the Hamiltonian constraint, and the ground-truth trajectory. We fix the training prediction horizon to $10$ frames and roll out the models to $30$ and $50$ frames. The results are summarized in Table~\ref{tab:overall_motion_results}. \method{} consistently yields lower physical-state prediction errors under both rollout horizons. Notably, the improvements are substantially larger for higher-order motion quantities. Under the $30$-frame rollout, \method{} reduces the acceleration RMSE from $10.966$ to $9.235$, corresponding to a $15.8\%$ reduction, and decreases the jerk RMSE from $14.135$ to $11.401$, yielding a $19.3\%$ improvement. The same trend persists under the longer $50$-frame rollout, where the acceleration and jerk errors are reduced by $10.2\%$ and $12.8\%$, respectively. These results demonstrate that Hamiltonian modeling improves the ability to capture instantaneous system dynamics and recover accurate motion states. We further visualize the predicted velocity trajectories in Fig.~\ref{fig:velocity_curve}. The velocity curves produced by \method{} exhibit substantially better agreement with the ground truth, whereas the non-Hamiltonian variant shows larger deviations and stronger fluctuations, further validating the effectiveness of the Hamiltonian constraint for instantaneous physical-state prediction.

\paragraph{Energy Conservation.}
Finally, we evaluate the physical consistency of generated trajectories from the perspective of energy conservation. For each rollout, we estimate the underlying physical states from the generated video and compute the corresponding Hamiltonian(total system energy). We then compare the temporal evolution of the Hamiltonian produced by \method{}, the non-Hamiltonian variant, and the ground truth. For conservative systems, such as projectile motion under uniform gravity, the total mechanical energy should remain approximately constant. Therefore, Hamiltonian drift provides a principled measure of long-horizon physical consistency. We further visualize both the Hamiltonian trajectories and the corresponding energy drift in Fig.~\ref{fig:energy_curve}. As shown in the figure, the Hamiltonian predicted by \method{} remains significantly closer to the ground-truth energy throughout the rollout, whereas the non-Hamiltonian variant exhibits substantially larger drift and oscillations. These results demonstrate that the proposed Hamiltonian formulation effectively preserves global physical invariants, leading to long-horizon dynamics that are more faithful to the underlying physical system.

\section{Real-World Large World-Model Experiments: Visual Generation Quality}
\label{app:real_world_visual}

We next evaluate \method{} as a drop-in transition module in real large world models. The key question here is whether a more physically grounded latent transition can improve not only physical plausibility, but also the standard visual generation quality used in practice to assess modern embodied world models.

\textbf{Ctrl-World.}
On Ctrl-World, which emphasizes controllable single-arm multi-view manipulation \citep{guo2026ctrlworldcontrollablegenerativeworld}, \method{} improves visual quality while simultaneously strengthening cross-view consistency. Replacing the generic transition reduces view-dependent drift and better preserves the correspondence between commanded actions and the motion observed from different cameras, yielding more stable open-loop video generation across viewpoints.

\textbf{EVAC.}
On EVAC \citep{jiang2025enerverse}, which targets dual-arm single-view manipulation, \method{} again improves visual generation quality over the original backbone. The gain is most apparent in coordinated bimanual interactions, where physically implausible drift in relative motion and contact timing is otherwise amplified over time. This result shows that the benefit of Hamiltonian structuring is not tied to a single observation layout or embodiment regime, but transfers across substantially different world-model families.

\textbf{Details.} We train our model on the large-scale real-robot dataset \textit{AgiBot World Challenge}, which encompasses 10 typical task categories across domestic, industrial, and medical scenarios. For evaluation, we adopt \textit{EWMBench} \citep{yue2025ewmbench} as the benchmark, which quantifies generation quality from five dimensions: appearance fidelity (PSNR, SSIM), temporal scene consistency (Scene), gripper trajectory alignment (nDTW), semantic description similarity (BLEU, CLIP), and task/physical logic (Logic). All fidelity metrics are higher-the-better. The test set of EWMBench is derived from held-out samples of AgiBot World, and the evaluation pipeline is fully automated to ensure reproducible results.

\section{Real-World Large World-Model Experiments: Task Execution Success Rate}
\label{app:cosmos_execution}

\begin{figure}[t!]
  \setlength{\abovecaptionskip}{-0.1cm}
  \setlength{\belowcaptionskip}{-0.2cm}
  \centering

    \includegraphics[width=\columnwidth]{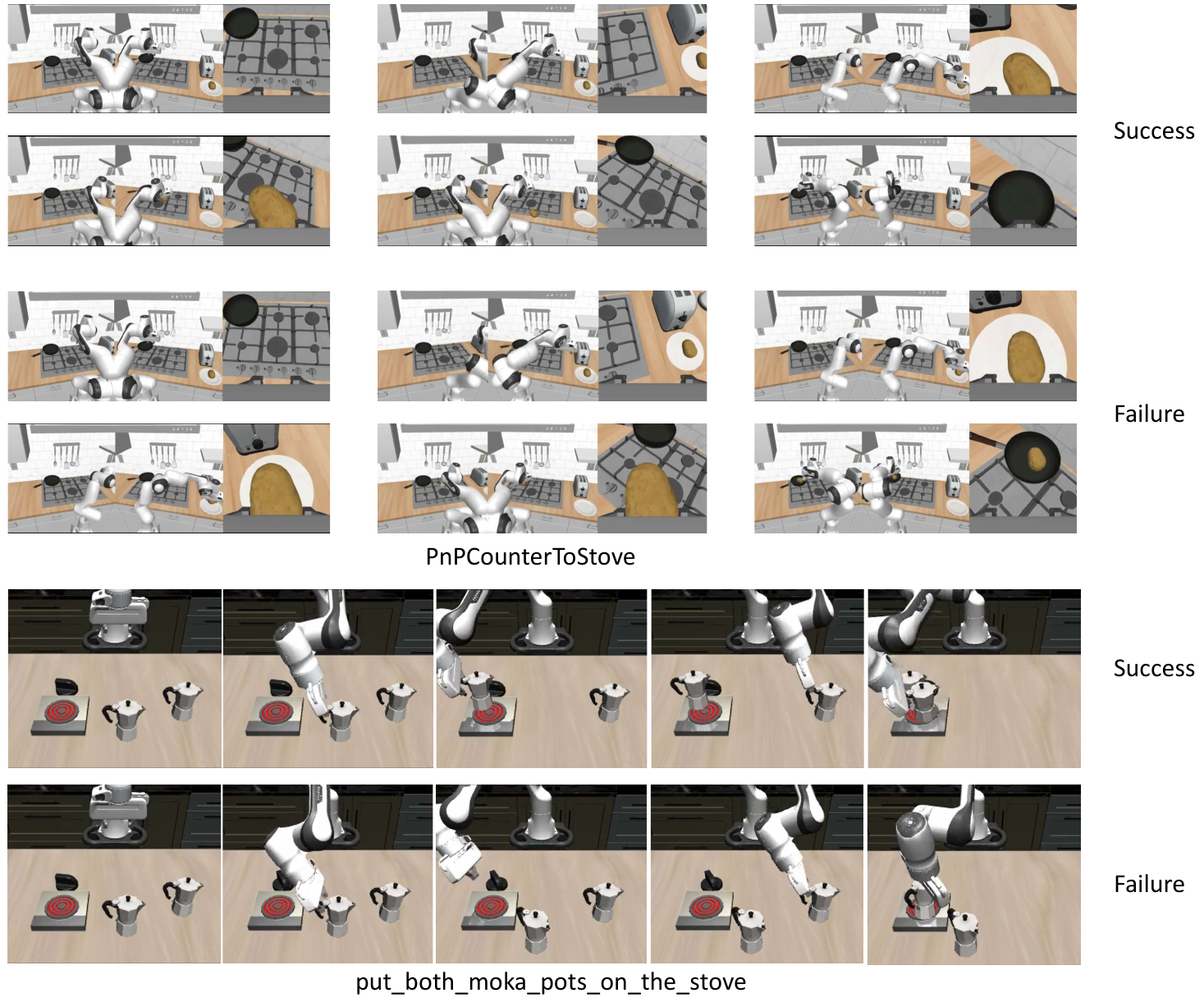}
  \caption{Examples of \method{} (successful) and Vanilla (failed) executions on two representative tasks: PnP (Pick-and-Place) and Put. The top rows show the PnP task (PnPCounterToStove), while the bottom rows show the Put task (put\_both\_moka\_pots\_on\_the\_stove). For each task, successful and failed trajectories are presented for comparison.}
  \label{fig:overview}
\end{figure}

\textbf{COSMOS.}
To evaluate whether the proposed Hamiltonian latent dynamics improve the practical usefulness of large world models for embodied decision making, we conduct task execution experiments with \textit{Cosmos-Predict2.5-2B} as the backbone world model. Rather than evaluating visual prediction quality in isolation, we measure whether imagined future trajectories remain sufficiently consistent with the underlying environment dynamics to support successful task execution. We evaluate \method{} on two widely used simulation benchmarks, LIBERO and Robocasa, which cover both long-horizon manipulation and diverse kitchen interaction tasks. As shown in Table~\ref{tab:overall_libero_robocasa}, \method{} consistently outperforms the vanilla model on both benchmarks. On LIBERO, our method achieves an average success rate of 98.7\%, improving over the vanilla model by 0.9 percentage points, with gains across all four task categories and the largest improvement on the Goal subset (from 96.2\% to 98.0\%). On the more challenging Robocasa benchmark, \method{} further improves the average success rate from 66.7\% to 68.5\%. These results indicate that the proposed Hamiltonian latent dynamics produce more reliable imagined trajectories, resulting in more effective downstream task execution. To further understand the source of these improvements, we provide qualitative case studies in Figure~\ref{fig:overview}, which compare successful executions of \method{} against failure cases of the vanilla model on two representative manipulation tasks.

\begin{table*}[t]
\centering
\caption{Task-level success rates across three seeds on COSMOS with Robocasa. V and H denote Vanilla and \method{}, respectively.}
\resizebox{\textwidth}{!}{
\begin{tabular}{c|ccc|c|ccc|c|c}
\hline
Task & V195 & V196 & V197 & V$_{avg}$ &
H195 & H196 & H197 & H$_{avg}$ & $\Delta$ \\
\hline
CloseDoubleDoor & 88 & 84 & 88 & 86.7 & 90 & 86 & 96 & 90.7 & +4.0 \\
CloseDrawer & 98 & 98 & 98 & 98.0 & 98 & 100 & 100 & 99.3 & +1.3 \\
CloseSingleDoor & 96 & 96 & 96 & 96.0 & 96 & 92 & 100 & 96.0 & 0 \\
CoffeePressButton & 98 & 96 & 98 & 97.3 & 96 & 94 & 100 & 96.7 & -0.7 \\
CoffeeServeMug & 58 & 52 & 48 & 52.7 & 56 & 58 & 62 & 58.7 & +6.0 \\
CoffeeSetupMug & 40 & 20 & 38 & 32.7 & 38 & 28 & 44 & 36.7 & +4.0 \\
OpenDoubleDoor & 84 & 92 & 90 & 88.7 & 80 & 88 & 86 & 84.7 & -4.0 \\
OpenDrawer & 92 & 96 & 76 & 88.0 & 94 & 96 & 100 & 96.7 & +8.7 \\
OpenSingleDoor & 80 & 88 & 90 & 86.0 & 82 & 90 & 88 & 86.7 & +0.7 \\
PnPCabToCounter & 28 & 38 & 30 & 32.0 & 30 & 30 & 36 & 32.0 & 0 \\
PnPCounterToCab & 62 & 58 & 50 & 56.7 & 60 & 58 & 66 & 61.3 & +4.7 \\
PnPCounterToMicrowave & 26 & 40 & 26 & 30.7 & 30 & 36 & 36 & 34.0 & +3.3 \\
PnPCounterToSink & 62 & 50 & 52 & 54.7 & 58 & 48 & 64 & 56.7 & +2.0 \\
PnPCounterToStove & 50 & 74 & 66 & 63.3 & 50 & 68 & 56 & 58.0 & -5.3 \\
PnPMicrowaveToCounter & 40 & 34 & 24 & 32.7 & 40 & 46 & 46 & 44.0 & +11.3 \\
PnPSinkToCounter & 74 & 70 & 70 & 71.3 & 70 & 76 & 76 & 74.0 & +2.7 \\
PnPStoveToCounter & 68 & 72 & 60 & 66.7 & 60 & 66 & 66 & 64.0 & -2.7 \\
TurnOffMicrowave & 100 & 100 & 100 & 100 & 100 & 100 & 100 & 100 & 0 \\
TurnOffSinkFaucet & 70 & 76 & 78 & 74.7 & 72 & 78 & 78 & 76.0 & +1.3 \\
TurnOffStove & 20 & 24 & 28 & 24.0 & 14 & 20 & 20 & 18.0 & -6.0 \\
TurnOnMicrowave & 76 & 76 & 84 & 78.7 & 76 & 82 & 82 & 80.0 & +1.3 \\
TurnOnSinkFaucet & 64 & 60 & 60 & 61.3 & 64 & 70 & 70 & 68.0 & +6.7 \\
TurnOnStove & 52 & 54 & 50 & 52.0 & 52 & 58 & 58 & 56.0 & +4.0 \\
TurnSinkSpout & 72 & 80 & 76 & 76.0 & 72 & 78 & 78 & 76.0 & 0 \\
\hline
\end{tabular}
}

\label{tab:task_seed}
\end{table*}

\textbf{LIBERO.}
We follow the standard LIBERO simulation benchmark protocol~\citep{liu2023libero}. LIBERO consists of a single Franka Emika Panda robot arm interacting with a variety of objects and environments, and provides four primary task suites: LIBERO-Spatial, LIBERO-Object, LIBERO-Goal, and LIBERO-Long. These suites evaluate different aspects of manipulation generalization, including spatial layout variation, object variation, language-specified goals, and long-horizon task composition. Each suite contains 10 tasks with 50 demonstrations per task, resulting in 500 demonstrations per task suite.

For our experiments, Cosmos-Predict2.5-2B is used as the base world model, and we compare the original transition model against \method{} while keeping the remaining policy and planning components fixed. Following the standard data processing protocol~\citep{kim2026cosmos}, unsuccessful demonstrations are filtered from the policy-training data, while the complete unfiltered demonstration set is retained for world-model and value-function training. We report task success rate across the four LIBERO suites, with particular emphasis on LIBERO-Long, where accumulated transition errors over extended action sequences provide a more demanding test of the learned dynamics.

The results show that incorporating Hamiltonian latent dynamics consistently improves task execution success over the original Cosmos transition. The gains are especially pronounced on long-horizon tasks, where small one-step prediction errors can otherwise compound into increasingly inaccurate imagined trajectories. By enforcing a structured evolution in latent phase space, \method{} produces more stable intermediate states and more reliable action-conditioned predictions, allowing the downstream policy to maintain a coherent sequence of actions for a longer duration. These results indicate that improvements in latent transition quality translate directly into better downstream manipulation performance rather than merely improving reconstruction fidelity.

\textbf{Robocasa.}
We further evaluate \method{} on the Robocasa simulation benchmark~\citep{nasiriany2024robocasa}, which contains 24 static kitchen manipulation tasks executed by a single Franka Emika Panda robot arm. Following the evaluation protocol of prior works~\citep{kim2026cosmos}, we evaluate each task over 50 trials distributed across five evaluation scenes with different floor plans and visual styles, with 10 trials per scene. The final success rate is averaged across all 24 tasks and three random seeds, corresponding to 3600 evaluation trials in total.

Unlike LIBERO, the Robocasa evaluation protocol emphasizes generalization to unseen object instances, and two of the five evaluation scenes for each task contain styles that are not observed during training. This setting therefore provides a stronger test of whether the learned world model captures transferable interaction dynamics rather than memorizing appearance-specific correlations.

To isolate the data efficiency of the proposed dynamics model, we train on the 50 human-teleoperated demonstrations available for each task and do not use the additional MimicGen demonstrations. As in the LIBERO experiments, unsuccessful demonstrations are removed for policy training, while the complete demonstration set is retained for world-model and value-function training. We then evaluate the task execution success rate under the standard Robocasa protocol.

Compared with the original Cosmos transition model, \method{} achieves higher average task success across the 24 Robocasa tasks. The improvement is particularly meaningful in scenes containing unseen styles and object instances, where purely appearance-driven prediction can be insufficient to maintain consistent object interactions over long action sequences. The structured Hamiltonian transition reduces accumulated latent drift and produces more coherent action-conditioned futures, which in turn improves the reliability of downstream planning and policy execution.

Overall, the LIBERO and Robocasa results demonstrate that the benefit of \method{} extends beyond improved latent or visual prediction metrics. When integrated into Cosmos-Predict2.5-2B, the proposed Hamiltonian transition consistently improves downstream task execution success, with the largest gains appearing in settings that require long-horizon reasoning, sustained object interaction, and generalization beyond the visual configurations encountered during training. These results provide direct evidence that physically structured latent dynamics can improve the usefulness of large world models for embodied control.

\end{document}